# FUZZY INFERENCE SYSTEM FOR INTEGRATED VVC IN ISOLATED POWER SYSTEMS

Vega-Fuentes E[1], Cerezo-Sánchez J M[2], León-del Rosario S[3], Vega-Martínez A[4]

Institute for Applied Microelectronics
Dept. of Electronics Engineering and Automatics
University of Las Palmas de Gran Canaria, Spain
[1]evega@diea.ulpgc.es, [2]jcerezo@diea.ulpgc.es, [3]sleon@diea.ulpgc.es,
[4]avega@diea.ulpgc.es

## ABSTRACT

*This paper presents a fuzzy inference system for integrated volt/var control (VVC) in distribution substations. The purpose is go forward to automation distribution applying conservation voltage reduction (CVR) in isolated power systems where control capabilities are limited. A fuzzy controller has been designed. Working as an on-line tool, it has been tested under real conditions and it has managed the operation during a whole day in a distribution substation. Within the limits of control capabilities of the system, the controller maintained successfully an acceptable voltage profile, power factor values over 0,98 and it has ostensibly improved the performance given by an optimal power flow based automation system. CVR savings during the test are evaluated and the aim to integrate it in the VVC is presented.*

## 1. INTRODUCTION

The deregulation has set the focus upon distribution networks. Since electricity market reform the distribution system operators have been subject to new regulations to secure an adequate level of reliability of supply and to improve the efficiency of the utility.

Conservation Voltage Reduction (CVR) is a known method for reducing system load and to save energy in a cost effective way. Based on the fact that certain loads can change with voltage.

Traditionally, the substation bus voltage is set at a high value in order to guarantee an adequate voltage level to customers at the end of the distribution lines at on-peak periods. Integrated Volt/Var control enables the application of CVR. Lowering voltages on the distribution system in a controlled manner can reduce peak demand, losses and achieve more energy savings while keeping the lowest customer utilization voltage consistent with levels determined by regulatory agencies and standard-settings organizations [1].

The purpose of Volt/Var Control (VVC) is the coordinated control of reactive power and voltage. VVC is a fundamental operating requirement of all electric distribution systems. It maintains acceptable voltage at all points along the distribution feeder under all loading conditions and restrains reactive power flow reducing the losses in transmission and subtransmission networks in electric power systems.

The control is achieved by on load tap changers (OLTC) in main transformers, by staggered shunt capacitor banks that can be located on the feeders or at the secondary bus of the substation and sometimes even by tie-switches on the feeders which reconfigure the network.

Optimal Power Flow (OPF) based automation systems are often found managing VVC [2], [3], [4], [5]. [6]. OPF improves an objective function, usually minimizing subtransmission losses. To deal with this, OPF compares the current status with that achievable by acting on control variables, if the calculated new state improves the objective function then outputs are acted.

The computational burden is high, due to power flow calculations for every secondary bus in all substations and the architecture of this solution must be centralized because the algorithm needs data from multiple locations. Local implementations at substations are discarded because of the huge amount of data exchange required.

Contrary to great quantity of works on transmission systems and distribution feeders, papers referred to distribution substation are rather limited [7], [8]. Reports on transmission systems [9], [10] ignore the singularities of isolated and small power systems where there is not subtransmission, voltage levels greater than 36 kV are treated as transmission and in deregulated market its operation is not distributor responsability. Shunt capacitor banks are not installed to work stepwise due to breakers costs. There are no shunt capacitors on the feeders, and large consumers have their own capacitors although the distributor has no way to manage their connection.

In point of fact limited size of these networks has led distribution network utilities to consider any specific development negligible, so the same applications as in continental power systems are used. This means that lot of functionalities do not apply and that the computational burden generates high response times.

In practice, operators at distribution and dispatch control center, with heuristic rules based on their past experience decide better strategies for voltage/reactive power regulation than these unsuitable programs. However as described in previous paragraphs, power system networks, strongly based on SCADA systems, are expected to evolve for Intelligent SCADA, and distribution automation as essential piece of smart distribution, is requested.

In this paper, a soft computing system based in fuzzy logic is proposed to deal with voltage/reactive power regulation in the Canary Islands (Spain) distribution substations. Its guiding principle, "exploit tolerance for imprecision, uncertainty and partial truth to achieve tractability, robustness and low costs solution" fits exactly with the purpose of the pursued system.

The controller has been implemented in the control center because of operative reasons for the pilot test, however the objective is to embed it in the substation control units at distribution substations.

This paper's composition is as next, section 2 describes the system under study, the control devices and their operation policy. Section 3 introduces optimal power flow applications for voltage and reactive power control problem and their inefficiencies and limitations when applied to isolated and small power systems and with the substation automation issue. Section 4 proposes a fuzzy control system to deal with the regulation task. The report of a pilot test carried out in a substation working under real conditions is presented. Section 5 focuses on CVR, as voltage profile has been flattened, lower voltage levels have supplied customers. The effects on demand curve and on energy savings are evaluated, and the achievable results by changing operation policy are estimated.

## 2. DESCRIPTION OF THE SYSTEM

The analyzed system is part of a 66/20 kV distribution substation as shown in Fig. 1, it is composed by a 66 kV primary bus, a 20 kV secondary bus and a 50 MVA power transformer equipped with OLTC to regulate the secondary bus voltage and keep it nearby its specified value under changing load conditions. This is accomplished with 6 taps under and 15 taps over nominal settings, with voltage steps of 1.46 %.

Located at the secondary bus of the substation there are shunt capacitor banks capable of providing 4.2 MVAr but connected through one only breaker, limiting the reactive power compensation to an all or nothing configuration. The operation policy is to limit the reactive power flow through the transfomer but avoiding recirculation, setting a maximum value of leading power factor beyond which the capacitive way of working of the transformer would produce undesirable effects on generation, specially when demand falls at off-peak hours.

Reactive power management policies remain insufficiently developed, mainly due to the local nature of the problem and the complexity of developing a competitive market for the provision of reactive power [11]. Thus, voltage and reactive power control is normally included in the category referred to as ancillary services, which are necessary for the efficient and economical provision of active power.

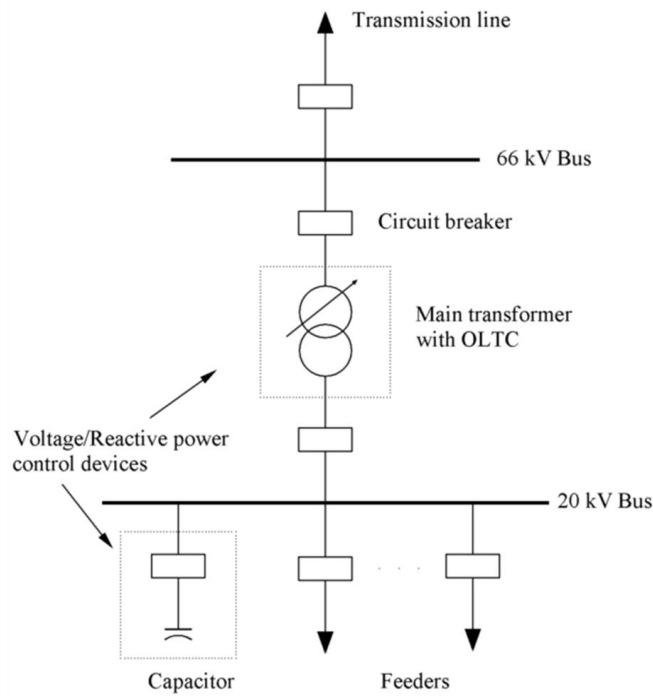

Figure 1. Part of a 66/20 kV distribution substation in the Canary Islands

The feeders connected to secondary bus include industrial and domestic customers. Constant impedance, constant current and constant power load models should be considered in order to describe the relationship between load demands and bus voltage. This distribution network has a slightly meshed topology but it is radially operated.

The SCADA system manages six stand alone electrical power networks, almost one per island, over 52 HV/MV substations and about 900 remotely controlled MV distribution centers. There are two frontends coexisting, the main one comunicates with remote terminal units (RTUs) using IEC 60870-5-104:2006 protocol. The other one use WISP+ extended protocol, installations which still communicate with this frontend are gradually migrating to the main one.

The RTUs are solved with substation control units which carry out the communications and data management of substation protection, control, and metering intelligent electronic devices (IEDs) using IEC 60870-5-101:2003 protocol. They also provide a local human-machine interface.

Despite the classes of precision of the metering transformers at substations are 0.2 – 0.5 for the variables involved in voltage/reactive power regulation and at IEDs metering input cards the accuracy is 0.1%, the measurement resolution at SCADA is only: 100 V, 10 kW, 10 kVAr and 1 tap position. This is an important matter because as the refresh time of the value is just 4 seconds it is common the observation of oscillations of hundreds of volts and one of the only two control outputs, the tap changer, has voltage steps of 1.46% that is about 320 V. So, the effects of data uncertainty should be taken into account.

## 3. OPTIMAL POWER FLOW FOR REACTIVE POWER/VOLTAGE CONTROL PROBLEM

OPF applications improve objective functions, usually minimizing subtransmission losses or flattening the voltage profile with the aim of attaining a better quality of service. To deal with this, OPF compares the current status with that achievable by acting on control variables, in this case connecting shunts or changing transformer taps. If the calculated new state improves the objective function then outputs are acted.

The computational burden is high, due to power flow calculations for every secondary bus in all HV/MV substations. The response time is high too, so the solutions to under or overvoltage are delayed except when limits are exceeded, in those cases warnings are activated and operators act. Furthermore, as oscillations of the metered values are not taken into account and response time is high, the current status at the moment of the result control action may be quite different than that used for calculations. It has been pointed that required data for power flow calculations exceed the quality available.

Other aims such as limitation of switching number of capacitor or OLTC in a day, daily scheduling, and load characterization are dismissed.

Power flow calculations need data from multiple locations so local implementations at substations are discarded because of the huge amount of data exchange required.

Fig. 2 shows the voltage profile at secondary bus of the substation under study during a random day. The power factor at the boundary between transmission and distribution is shown in Fig. 3.

The desired value for voltage at secondary bus is 21.0 kV and the operation limits are defined at 20.3 and 21.6 kV. At those points the system alerts operators with low or high voltage alarms. It may be observed that although the profile shows the voltage inside operation zone, there are peaks during the afternoon and during the night. A better OLTC performance should be desirable. It acted 8 times keeping the tap position 2 from 23:23:38h even with 21,5 kV till 07:05:25h in the morning when voltage dropped to 20.9 kV.

Nowadays there is not a specific limit for power factor although in order to reduce losses in transmission network and to improve power transformer performance, values over 0.98 are desired. The profile shows leading power factor values close to 0.93 at 03:00 am and at 04:00 am. From 23:00 pm to 08:00 am the current is leading the voltage making the power transformer work in a capacitive way with its adverse associated effects. Despite of this, capacitor banks remained connected all day long. This illustrates the bad behavior of the running OPF based automatism.

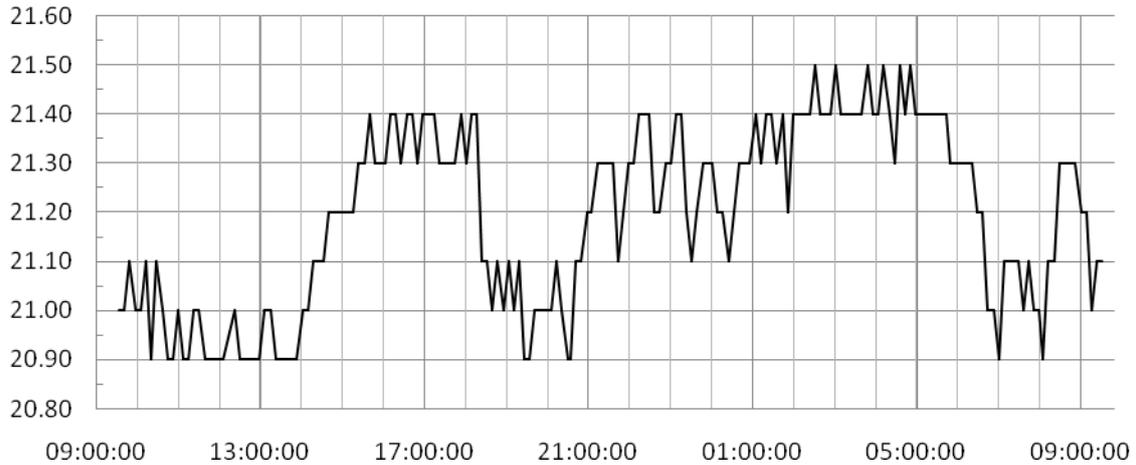

Figure 2. Voltage profile (y axis, kV) at secondary bus working with OPF

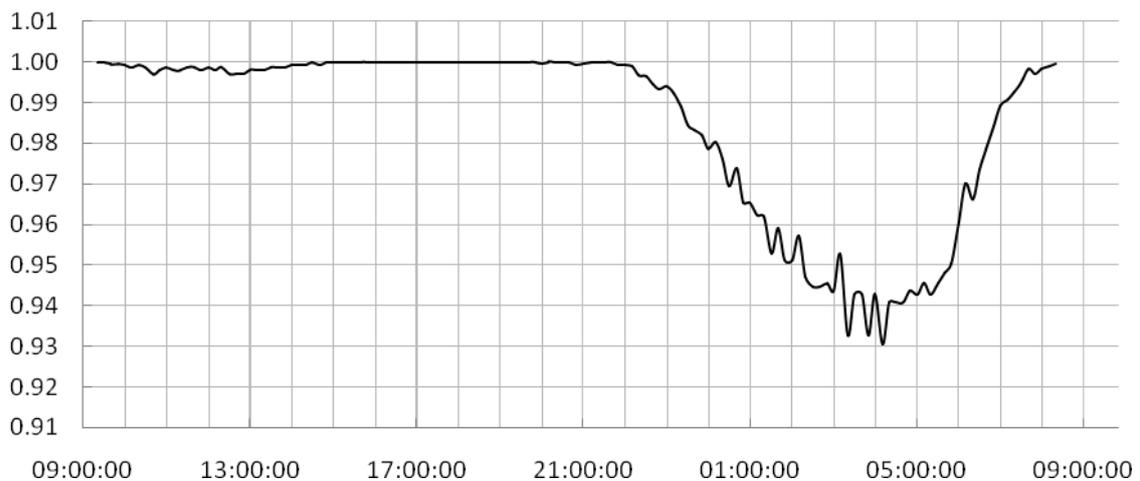

Figure 3. Power factor profile (y axis) at HV/MV boundary working with OPF

In paradigm of deregulated markets, the System Operator manages reactive power through three aspects applicable to all agents connected to the transmission network which are generators, transmission service providers and consumers:

- Reactive power constraints.

- Reactive power actually demanded.

- Payment for the service (penalties and incentives).

Distribution networks utilities are considered to be large consumers, this means that connection conditions in form of power factor constraint bands will be defined at every boundary transformer so reactive power profiles like shown would be penalized.

## 4. FUZZY LOGIC FOR REACTIVE POWER/VOLTAGE CONTROL PROBLEM

In point of fact systems such as this under study has few control options for reactive power/voltage control, so any trained operator would solve the task at least as fine as the OPF, with a lower response time and surely in a better way because furthermore from the regulation aim, other issues would be regarded such as time of day and expected increase of load, which leads to keep the voltage at the secondary bus a little bit over nominal values, the switching number of shunts and OLTC, issues to take into account in order to extend their life time, as well as imprecision, uncertainties and oscillations given by sensors, transmitters, acquisition systems and its digitalization process that show a partial truth of reality.

The way proposed to transmit operators knowledge and their heuristic rules to deal with these aims to an automatic system is through fuzzy logic, a tool for embedding structured human reasoning into workable algorithms.

Many fuzzy inference systems has been proposed for voltage/reactive power control [12], [13], [14], [15] and even adaptive neuro-fuzzy inference systems [16] which provide a method for the fuzzy modeling procedure to learn information about the data set, in order to compute the membership function parameters that best allow the associated fuzzy inference system to track the given input/output data. However, the fuzzy logic in these cases has been oriented to:

- Estimation of sensitivities of load profiles.

- Find a feasible solution set to achieve buses voltage improvement and then identify the particular solution which most effectively reduces the power loss.

- Manage reactive power resources in a transmission network.

- Optimize an objective function which includes fuzzified variables based on a forecast of real and reactive power demands.

But always in a loop with a power flow routine that evaluates the progressive effect of control action until a criterion is met. These algorithms probably improve the performance given by the OPF based control system but preserve all disadvantages that disadvise it for isolated and small power systems. On the other hand, as commented in previous section, power flow calculations and its associated data traffic do not fit well with substation automation task.

A fuzzy inference system (FIS) has been designed as an on-line, real time tool, to give the proper dispatching strategy for capacitor switchings and tap movements such that satisfactory secondary bus voltage profile and main transformer power factor are reached.

The fuzzy controller acts directly from the result of its inference system. No further calculations are executed in order to estimate the future status achievable and to compare it to the current one. As previously mentioned, it is assumed that operators at distribution and dispatch control center, may decide good strategies for voltage/reactive power regulation within the limits of the control capabilities of the system. So the work done consisted in the transmission of this knowledge from operators to the controller through heuristic rules based on their past experience, allowing it working in automatic way.

The input variables defined were: voltage at secondary bus, reactive power flow through the transformer measured at HV winding, tap position and shunt capacitor status switched on or off. Fig. 4 shows the fuzzy model membership function for the input Voltage.

Where the fuzzy linguistic sets for the input Voltage were: very low (VL), low (L), low on-peak time period (LP), good (G), high (H), high on-peak time period (HP) and very high (VH).

The ouput variables defined were orders to move up or down the OLTC and switch on or off the shunt capacitor banks. Fig. 5 shows the fuzzy model membership function for the output Taps.

The inference system was designed using Matlab, it was Mamdani type with 14 rules. The logic guidelines embedded were like:

- If (Reactive_power is High) and (Tap is Normal) and (Shunt_Off is Disconnected) then (Tap is -2)(Capacitor is Connect).

- If (Voltage is H) and (Reactive_power is Good) and (Tap is not Tap1) then (Tap is -1).

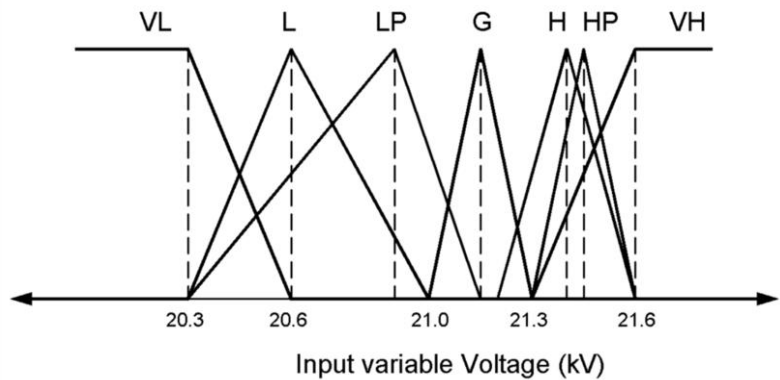

Figure 4. Sample of the fuzzy model membership functions. Input Voltage

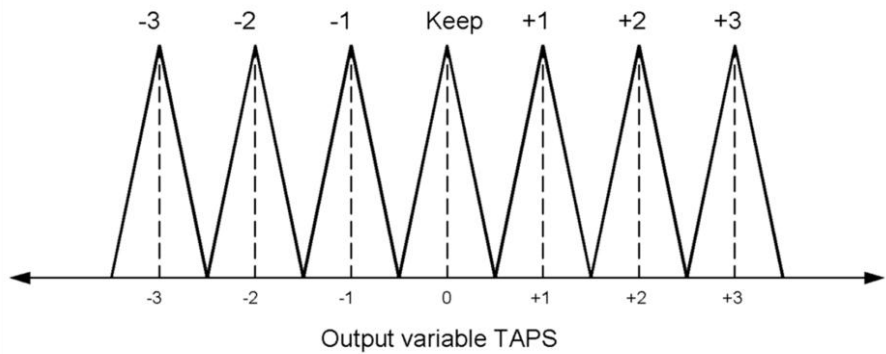

Figure 5. Sample of the fuzzy model membership functions. Output Taps

Fig. 6 shows the voltage profile at secondary bus of the substation and Fig. 7 shows the power factor at the boundary between transmission and distribution registered one week later than those showed working with OPF, this time working with the designed FIS.

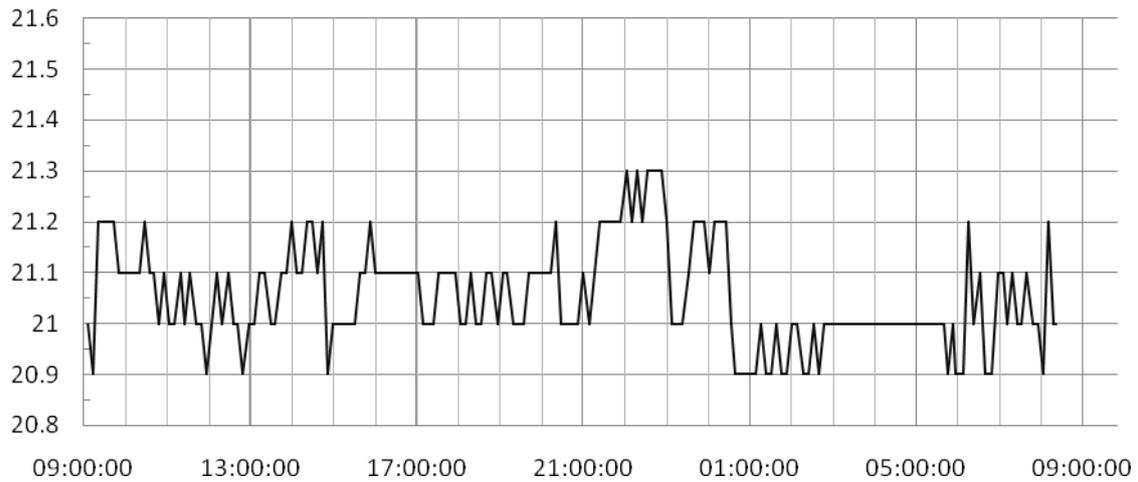

Figure 6. Voltage profile (y axis, kV) at secondary bus working with FIS

The voltage profile has been flattened and the power factor has been kept 23 hours of the day over 0.99 and always over 0.98. The OLTC just acted 11 times, a value well bellow the maximum defined (the maximum switching number of shunts and OLTC were defined as 30 and 6 respectively in order to extend their life time). The capacitor banks were disconnected from the secondary bus at 23:55:30h and were reconnected again at 08:13:19h.

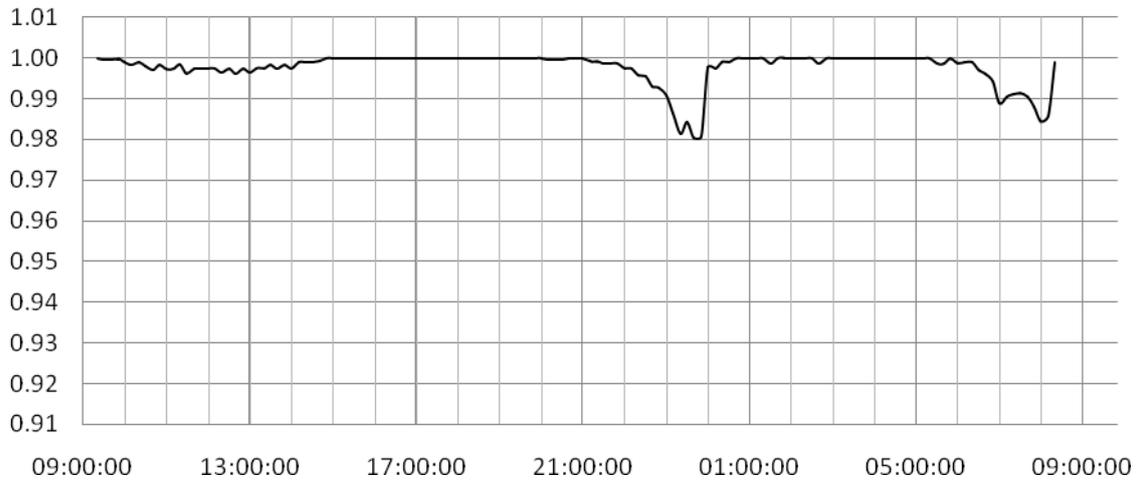

Figure 7. Power factor profile (y axis) at HV/MV boundary working with FIS

Fig. 8 shows the voltage profile at secondary bus comparation between two random days (april 16 and 22) working with an OPF based controller and one day (april 23) working with a FIS based controller.

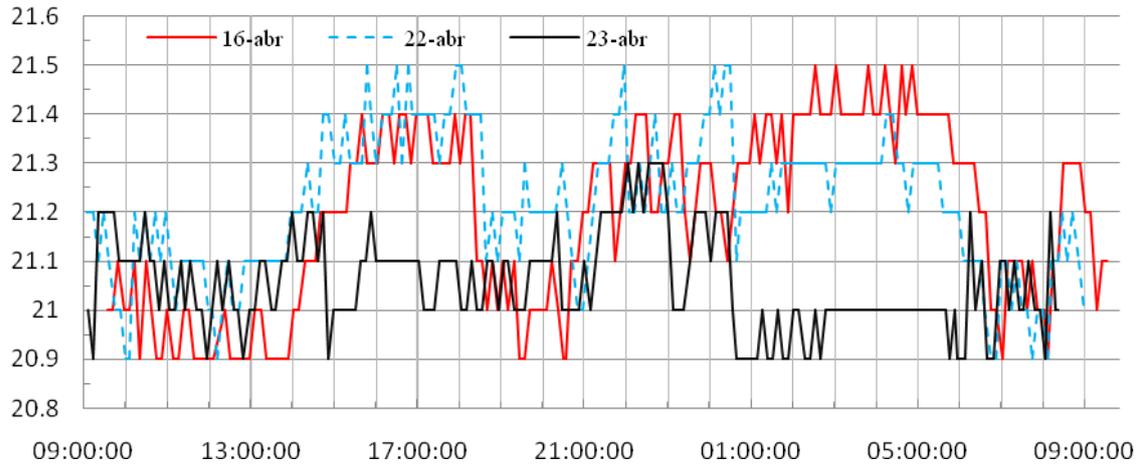

Figure 8. Voltage profile (y axis, kV) at secondary bus comparation

Table 1 shows statistical summary related to these voltage profiles. Average (Ū), maximum deviation (DM) and mean deviation (Dm) both refered to objective value calculated using expressions (1) and (2).

$$D_M = Max \ |U_i - 21| \tag{1}$$

$$D_m = \frac{\sum_{i=1}^{n} |U_i - 21|}{n} \tag{2}$$

where $U_i$ is the voltage value at the measure i, and n is the number of registered measurements.

Table 1. Statistical summary of voltage profiles results.

|   | **OPF** (Apr-16) | **OPF** (Apr-22) | **FIS** (Apr- 23) |
|---|---|---|---|
| **Ū** | 21.1914 kV | 21.2178 kV | 21.0537 kV |
| **D_M** | 0.5000 kV | 0.5000 kV | 0.3000 kV |
| **D_m** | 0.2192 kV | 0.2251 kV | 0.0792 kV |

The average voltage has been kept exactly in the reference value considering its measurement resolution. The maximum deviation has been reduced 200 V and the mean deviation had a 64 % drop off.

In order to evaluate the benefits of reactive power compensation, losses at transmission network with the system working with OPF based automatic control and with the FIS based one are compared. For this purpose only Joule effect losses are taken into account as they are the most important ones. The relationship is evaluated by expression (3).

$$\phi = \frac{Losses_{FIS}}{Losses_{OPF}} = \left(\frac{\cos\varphi_{OPF}}{\cos\varphi_{FIS}}\right)^2 \qquad (3)$$

As power factor does not remain constant in either case, neither does the losses relationship. Its profile is shown in Fig. 9. Although there are values over 1, mostly at instants after connection and before disconnection of the capacitor banks due to the limit reactive power recirculation policy embedded in the FIS, values of 0.8660 are reached. It implies 13.40 % losses reduction.

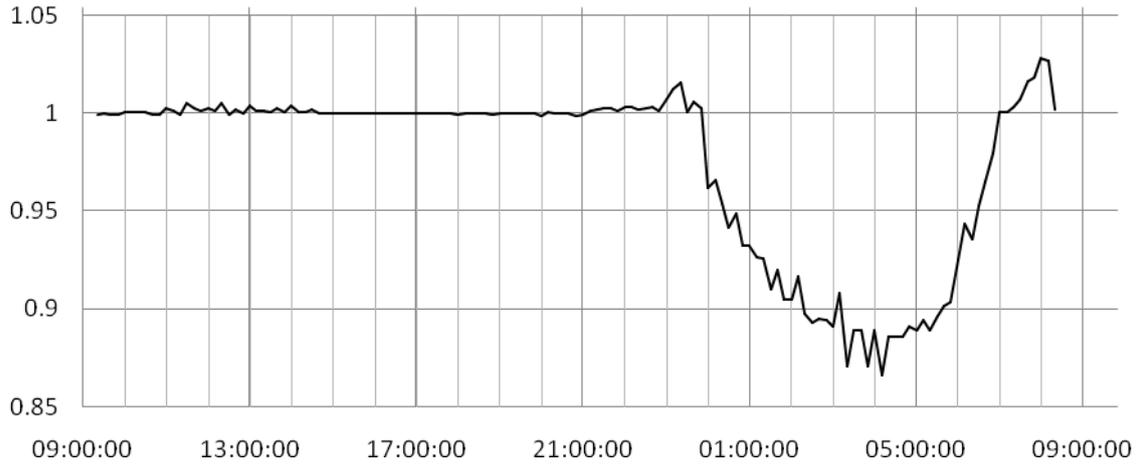

Figure 9. Losses ratio profile. FIS/OPF

The average losses ratio has been calculated using (4).

$$\overline{\phi} = \frac{\sum_{i=1}^{n}\left(\frac{\cos\varphi_{i\ OPF}}{\cos\varphi_{i\ FIS}}\right)^2}{n} \qquad (4)$$

where $\cos\varphi_i$ is the power factor value at the measure i, and n is the number of registered measurements.

Table 2 shows average losses ratio results for one day period comparation and during the interval during which FIS based system decided to disconnect capacitor banks and OPF based did not.

Table 2. Average losses ratio between OPF and FIS based controllers.

|  | $\overline{\phi}$ | **Loss Reduction** |
|---|---|---|
| **24h** | 0.9750 | 2.50 % |
| **23:55:30h to 08:13:19h** | 0.9312 | 6.88 % |

The average loss reduction for all day comparation was 2.50 % and in the interval during which controllers had differrent behavior the average loss reduction was 6.88 %. In order to achieve acceptable and valid results, days with the same load profile were compared. For this purpose it was selected the same day of the week before the day the FIS based controller was tested.

A better performance could be affordable with the capacitor banks working with several switches with staggered configuration however the results obtained ostensibly improve those achieved with the OPF based automatic control.

## 5. CONSERVATION VOLTAGE REDUCTION

CVR, also called conservative voltage regulation or voltage optimisation, is not a new methodology for energy conservation. There are papers that assess how much load reduction is gained by voltage reduction since 1977 [17]. Many loads operate more efficiently at lower voltage. With voltage optimization, we are optimizing voltages to loads, so that they operate as efficiently as possible with minimum disruptions.

The adoption of smart grid technologies, an all encompassing term reflecting the broad objective of applying the latest technology to the overall power system [18], it suggests greater efficiencies than ever before. It is expected that advanced smart metering will be available and giving feedback at every dispatch control center soon. This closed loop will enable adjusting the voltage at feeders to reduce the level at consumer terminals to the lowest regulated values. Meanwhile, integrated VVC, which is an important feature of the future grid, enables the application of CVR.

Spanish regulation [19] sets the range for voltages at final consumers terminals at 230 Volts ±7%. European Standard EN 50160 [20] indicates a variation range of the supply voltage of nominal voltage ±10%. American Standard C84.1 [21] defines 120 Volts ±5%. CVR is based on the principle that acceptable voltage band can be easily and inexpensively operated in the lower half (230-214 Volts in spanish case), without causing any damage to consumer appliances.

A way to explain the relationship between energy consumption an voltage is as follows. In a resistive circuit, the power P, voltage V and current I, satisfy Joule's law $P=VI$. When the load consists of pure resistors with constant resistance R, from Ohm's law $V = IR$. So for pure resistive load $P = V^2/R$ and lowering the voltage level reduces the power. Examples of constant resistance loads could be incandescent lighting, oven, microwave, etc.

Some lighting technologies (compact fluorescent lighting) keep the current constant. For these devices, the effect of CVR is reducing linearly the energy consumption.

For constant energy loads, these are constant resistance loads with a feedback loop which extend their operation cycle when submitted to undervoltages, the energy consumption is constant but distributed through a longer cycle, with a shaving effect over the demand curve. A good example of constant energy load is the electric water heater.

Finally, for constant power loads (computer, TV, etc.), CVR produces a raise of energy consumption, because of increasing line losses due to increase currrent draw caused by the lowered voltage. However this raising is negligible compared to savings obtained in constant impedance loads.

There are reports which categorise residential devices into constant-impedance, constant-energy and constant-power devices [22]. Fig. 10 shows the breakdown of the 2007 australian residential electricity consumption by load categories, where 'unknown' consists of a group of miscellaneous devices.

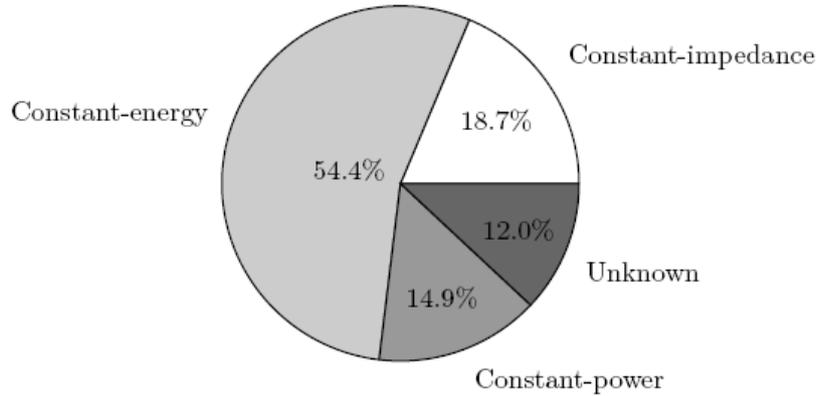

Figure 10. Breakdown of the 2007 australian residential electricity consumption

CVR is not only efficient for consumers. From utilities point of view, applying CVR can reduce the core losses in transformers including eddy current and hysteresis losses. As far as transmission line losses are concerned, the line current may increase slightly when voltage is reduced for constant power loads. Therefore the net system losses are reduced.

CVR effects can be evaluated by the conservation voltage regulation factor (CVR$f$), which is defined as follows:

$$CVRf = \frac{\Delta E \%}{\Delta V \%} \tag{5}$$

where $\Delta E\%$ is the percentage of energy reduction and $\Delta V\%$ is the percent voltage reduction. The ratio can also be calculated for active power o reactive power reduction, so the CVR$f$ is followed by kWh, kW o kVAr depending on what is the reduction referred to.

The average values for CVR factor assessed in field trials of voltage reductin on nine distribution circuits [23] according to load characterization are:

- Domestic customers. CVR$f$ (kWh) = 0.76
- Commercial customers. CVR$f$ (kWh) = 0.99
- Industrial customers. CVR$f$ (kWh) = 0.41

The mean values for the average feeder obtained through these field trials are:

- CVRf (kWh) = 0.69
- CVRf (kW) = 0.78
- CVRf (kWh) = 3.45

During the test carried out in this work, the average percent voltage reductions obtained from the voltage profiles statistical results shown in Table 1 were 0.65% and 0.77% referred to one week and one day before the test respectively. Applying the CVR factors, the expected saving values were: 0.44-0.53% kWh, 0.51-0.60% kW and 2.24-2.65% kVAr.

Fig. 11 shows the active power profile comparation at HV/MV boundary transformer measured at MV windings. The day operating with FIS presents a shaved demand curve, mostly on peak hours.

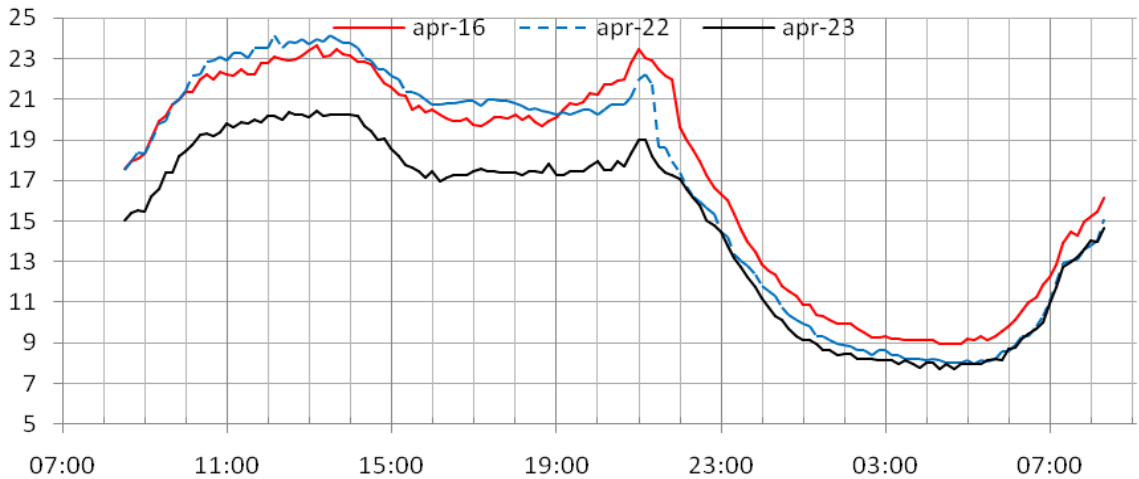

Figure 11. Active power profile (y axis, MW) comparation

Fig. 12 shows the reactive power profile comparison at HV/MV boundary transformer but this time measured at HV windings due to better quality of data. The profile working with FIS improves that one resulting from the system working with OPF, always within the control capabilities. Staggered capacitor banks would allow a better response, via a step by step connection or disconnection. It should also be desireble to increase the reactive power capacitor size for compensation at on peak hours.

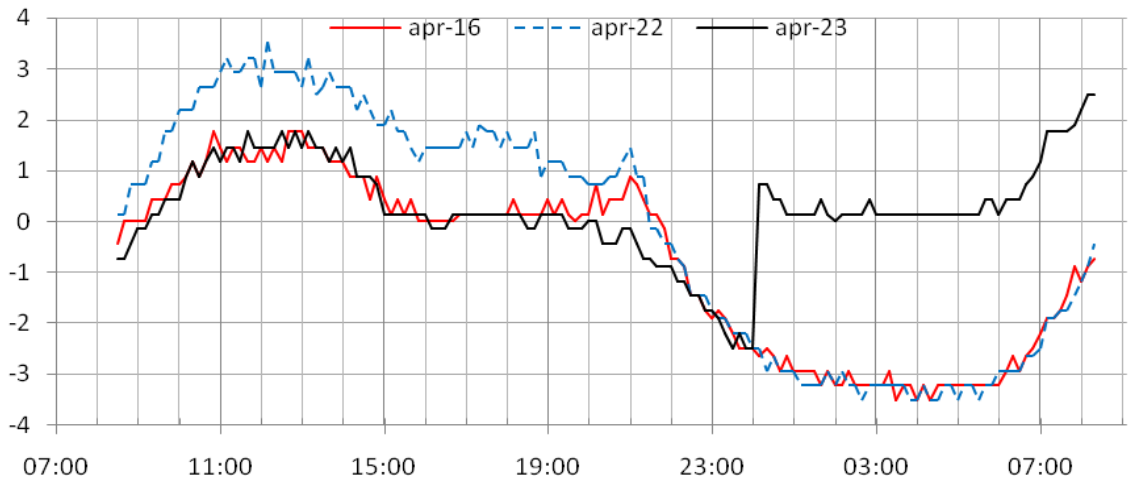

Figure 12. Reactive power profile (y axis, MVAr) comparation

Table 4 shows the statistical summary of power profiles results. Savings exceed the values expected. It is supposed that other factors should be taken into account so an specific test should be designed to evaluate the effects of CVR. Moreover, without real time feedback about end of the line voltage drops, a coservative value of voltage reduction should be considered, typically half the lower range (-2.5% for spanish regulation).

So to apply an integrated VVC enabling CVR, the operation policy for voltage would be modified and instead of 21.0 KV the new desired value at secondary bus should be 20.475 KV (21.0 kV– 2.5%). Applying the CVR factors again, the new expected saving values after the operation policy change are: 1.72% kWh, 1.95% kW and 8.62% kVAr.

Table 4. Statistical summary of power profiles results.

|  | **OPF** (Apr-16) | **OPF** (Apr-22) | **FIS** (Apr- 23) |
|---|---|---|---|
| $\bar{P}$ | 17.2075 MW | 16.7620 MW | 14.8236 MW |
| $\bar{Q}$ | -0.7811 MVAr | -0.1583 MVAr | 0.2622 MVAr |

## 6. CONCLUSIONS

A fuzzy inference system has been designed as an on-line, real time tool, to give the proper dispatching strategy for capacitor switchings and tap movements for control of reactive power/voltage in a distribution substation.

The system passed all the security matters around power systems and was tested in a real case under real conditions. It managed the operation at dispatch control center during a whole day for one distribution substation in the Canary Islands.

Within the limits of the control capabilities of the system, the fuzzy controller maintained successfully an acceptable and flattened voltage profile, reducing 36.13% (from 219.2V to 79.2V) the mean deviation refered to objective value. The reduction of computational burden has been significative and has led to lower time response when submitted to changing operation conditions. The average voltage has been kept exactly in the reference value considering its measurement resolution.

The power factor values are kept over 0.98 and has ostensibly improved the performance given by an optimal power flow based automation system. The ratio of losses improvement during a day period has been estimated in 2.50 % reaching 13.40 % loss reduction at certain times of the day .

The effects of conservation voltage reduction have been assessed. The savings obtained exceed widely those expected by applying CVR factors resulting from other studies. Further analisys including other factors should be done. Although an accurate CVR*f* may not be inferred, results denote efficiency.

The operation policy for voltage at substation secondary buses should be modified to attend conservation policy. The designed fuzzy controller could be reconfigured to keep new voltage reference.

It has been demonstrated that for distribution substations in stand alone electric power systems with limited control actions, fuzzy inference systems solve properly the integrated voltage/reactive power control task.

The logic proposed may be embedded in the substation control unit. This would release the control centre of these tasks, although retaining the ability to monitor, supervise and even configure the inference system. It should be taken into account in the challenge of substation automation in isolated and small power systems.

## SYMBOLS

$\bar{U}$:     Average voltage.
$D_M$ :     Maximum voltage deviation.
$D_m$ :     Mean voltage deviation.

$\phi$ : Losses ratio.
$\bar{\phi}$ : Average losses ratio.
CVR$f$ : Conservation voltage reduction factor.
$\bar{P}$ : Average power.
$\bar{Q}$ : Average reactive power.

## ACKNOWLEDGEMENTS

The authors would like to express their sincere gratitude to Endesa Distribución Eléctrica, S. L. and its staff at dispatch control center in Canary Islands for providing the valuable system data and dispatch experience.